\documentclass[runningheads]{llncs}

\usepackage{eccv}

\usepackage{eccvabbrv}

\usepackage{graphicx}
\usepackage{booktabs}
\usepackage[accsupp]{axessibility}  

\usepackage[pagebackref,breaklinks,colorlinks, citecolor=eccvblue]{hyperref}

\usepackage{orcidlink}

\usepackage{multirow}
\usepackage{indentfirst}

\def\0{{\bf 0}}
\def\1{{\bf 1}}

\definecolor{red}{rgb}{0.95,0.4,0.4}
\definecolor{purered}{rgb}{1,0,0}
\definecolor{blue}{rgb}{0.4,0.4,0.95}
\definecolor{darkblue}{rgb}{0,0,0.8}
\definecolor{darkred}{rgb}{1,0,0}
\definecolor{darkgreen}{rgb}{0,0.5,0}
\definecolor{grey}{rgb}{0.6,0.6,0.6}
\definecolor{col1}{RGB}{232, 161, 148}
\definecolor{col2}{RGB}{148, 187, 232}
\definecolor{lightgrey}{rgb}{0.85,0.85,0.85}
\definecolor{lightlightgrey}{rgb}{0.9,0.9,0.9}
\definecolor{verylightBG}{rgb}{0.9,0.99,0.99}
\definecolor{darkgreen}{rgb}{0.3, 0.75, 0.3}

\begin{document}

\title{
CriSp: Leveraging Tread Depth Maps for Enhanced Crime-Scene Shoeprint Matching
\vspace{-3mm}
}

\titlerunning{CriSp: Leveraging Tread Depth Maps for Crime-Scene Shoeprint Matching}


\authorrunning{Shafique et al.}

\author{
Samia Shafique\textsuperscript{1} \quad\quad
Shu Kong\textsuperscript{2,3,4\thanks{Authors share senior authorship.}} \quad\quad
Charless Fowlkes\textsuperscript{1$^{\star}$}
}

\authorrunning{S. Shafique et al.}

\institute{
\vspace{-1mm}
\textsuperscript{1}University of California, Irvine \quad\quad  
\textsuperscript{2}University of Macau
\\
\textsuperscript{3}Institute of Collaborative Innovation
\quad\quad     \textsuperscript{4}Texas A\&M University
\\
{\small {\tt sshafiqu@uci.edu} \quad 
{\tt skong@um.edu.mo}  \quad
{\tt fowlkes@ics.uci.edu}}
\\
\vspace{1mm}
{\small {\em code and dataset at} \url{https://github.com/Samia067/CriSp}}
\vspace{-7mm}
}

\maketitle

\begin{abstract}

Shoeprints are a common type of evidence found at crime scenes and are used regularly in forensic investigations. However, existing methods cannot effectively employ deep learning techniques to match noisy and occluded crime-scene shoeprints to a shoe database due to a lack of training data. Moreover, all existing methods match crime-scene shoeprints to clean reference prints, yet our analysis shows matching to more informative tread depth maps yields better retrieval results. The matching task is further complicated by the necessity to identify similarities only in corresponding regions (heels, toes, etc) of prints and shoe treads. To overcome these challenges, we leverage shoe tread images from online retailers and utilize an off-the-shelf predictor to estimate depth maps and clean prints. Our method, named \emph{CriSp}, matches crime-scene shoeprints to tread depth maps by training on this data. \emph{CriSp} incorporates data augmentation to simulate crime-scene shoeprints, an encoder to learn spatially-aware features, and a masking module to ensure only visible regions of crime-scene prints affect retrieval results. To validate our approach, we introduce two validation sets by reprocessing existing datasets of crime-scene shoeprints and establish a benchmarking protocol for comparison. On this benchmark, \emph{CriSp} significantly outperforms state-of-the-art methods in both automated shoeprint matching and image retrieval tailored to this task.

\end{abstract}

\section{Introduction}
\label{sec:intro}


\begin{figure*}[t]
  \centering
  \includegraphics[trim={1cm 10.2cm 5.9cm 2.2cm},clip,width=0.9\linewidth]{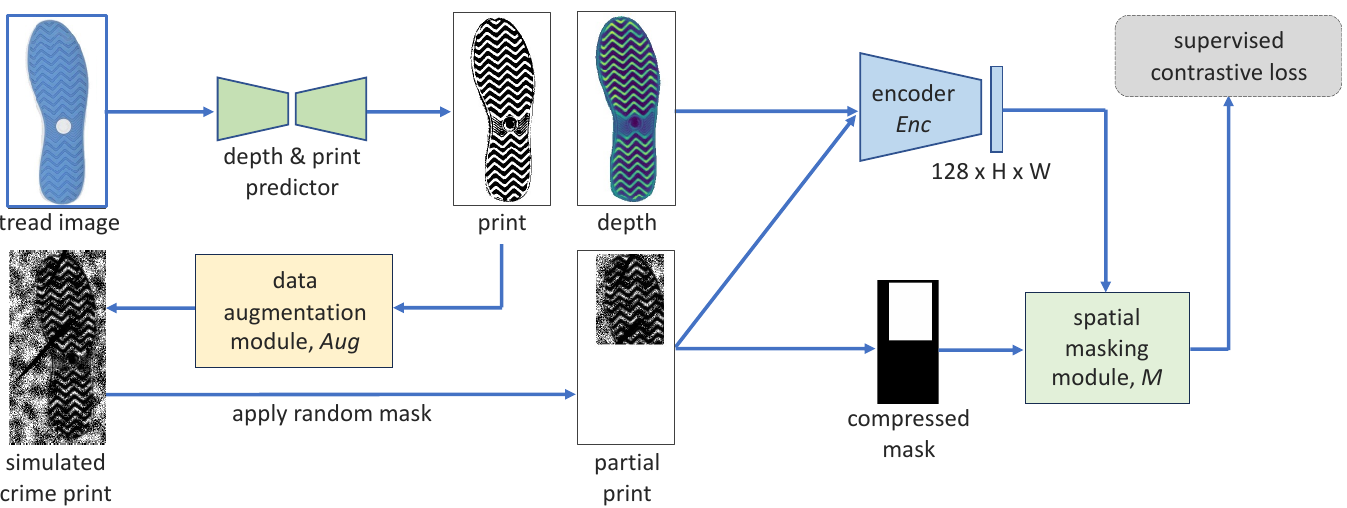}
  \vspace{-3mm}
  \caption{\small
  Our method \emph{CriSp} compares crime-scene shoeprints against a database of tread depth maps (predicted from tread images available at online retailers) and retrieves a ranked list of matches.
  We train \emph{CriSp} using tread depth maps and clean prints (\cref{sec:datasets}).
  We use a data augmentation module $Aug$ to address the domain gap between clean and crime-scene prints, and a spatial feature masking strategy (via spatial encoder $Enc$ and masking module $M$) to match shoeprint patterns to corresponding locations on tread depth maps (\cref{sec:methodology}). 
  \emph{CriSp} significantly outperforms previous methods (\cref{sec:experiments}).
  }
\label{fig:training_overview}
\vspace{-7mm}
\end{figure*}


Examining the evidence found at a crime scene assists investigators in identifying suspects. Shoeprints are likely to be found at crime scenes, despite their  fewer distinct identifying features than other biometric samples like blood or hair~\cite{bodziak2017footwear}. 
Hence, analyzing shoeprints can help criminal justice and forensics.

Examining shoeprints forensically offers insights into the class attributes and the acquired attributes of the suspect's footwear.
Class attributes pertain to the general features of the shoe, e.g., brand, model, and size.
Acquired attributes encompass the unique traits delivered by the shoe with wear and tear, e.g., holes, cuts, and scratches.
Our focus lies in facilitating the investigation of the class attributes of shoeprints.


{\bf Status quo.}
Traditional automated shoeprint matching methods \cite{bouridane2000application, algarni2008novel, wei2014alignment, kong2014novel, de2005automated, gueham2007automatic, kortylewski2015unsupervised, alizadeh2017automatic, almaadeed2015partial} typically use handcrafted features to match crime-scene shoeprints with clean, reference impressions. Recent ones \cite{zhang2017adapting, kong2019cross, ma2019shoe} use more generalizable features extracted by deep Convolutional Neural Networks (CNNs), which are usually pretrained on ImageNet \cite{deng2009imagenet} as  available shoeprint datasets \cite{fid300} are too small to train deep features.
This solicits large-scale shoeprint datasets for better solving the shoeprint matching problem.
Moreover, while existing methods match crime-scene shoeprints to clean reference shoeprints, we find that matching to tread \emph{depth maps} leads to significantly better performance 
(cf. \cref{table:data_augmentation}).

{\bf Motivation.}
To address the need for a large-scale training dataset, 
we leverage the extensive collection of tread images of various shoe products available at online retailers.
We generate tread depth maps and clearly visible prints using the method propsoed in \cite{shafique2022shoerinsics}. 
\cref{fig:training_dataset} shows some examples in our dataset.
Note that matching directly to RGB tread images causes models to overfit to irrelevant details such as albedo and lighting (\cref{subsection:design_choices}). Therefore,
\emph{we formulate our problem as the retrieval of tread depth maps that best match crime-scene shoerpints by learning a representation from tread depth maps and clean shoeprints.}

{\bf Technical insights.}
We develop a method termed \emph{CriSp} to address this problem using three key components (\cref{fig:training_overview}). 
First, a data augmentation module $Aug$  simulates crime-scene shoeprints from the clearly visible prints and depth maps of the training set. 
This helps mitigate domain gaps between our training set and real-world crime-scene testing images. 
Second, a spatial encoder $Enc$ ensures that our model learns to match patterns in corresponding regions of shoe treads.
For instance, if a crime-scene shoeprint exhibits stripes on the heels, the model must retrieve shoes with stripes in the heel region rather than other areas like the toe region. 
Third, a feature masking module $M$ ensures using only the visible parts of crime-scene shoeprints for retrieval. 
Our extensive experiments show that combining these components facilitates feature learning and yields significantly improved retrieval performance over prior arts.

{\bf Contributions.}
We make three major contributions:
\begin{itemize}
\item We introduce the concept of matching crime-scene shoeprints to tread depth maps, 
aiming to facilitate forensic investigation and criminal justice.
\item We propose a new benchmark consisting of a new dataset and retrieval-based evaluation protocols, allowing fair comparisons against previous methods.
\item We develop a spatially-aware matching method \emph{CriSp}, yielding superior performance over existing methods.
\end{itemize}


\section{Related Work}
\label{sec:related_work}

{\bf Automated shoeprint matching.}
The success of automated fingerprint identification systems~\cite{datta2001advances} has inspired the study of automated shoeprint matching~\cite{wu2022crime, li2021shoeprint,wen2023shoeprint}. 
Current literature aims to extract features from crime-scene shoeprints and match them to a database of laboratory footwear impressions to identify the shoe make and model~\cite{rida2019forensic}. 
Holistic methods process the shoeprint image as a whole,
e.g., reconstructing the shoeprint \cite{hassan2024deep}, and
representing shoeprints using Hu’s moment~\cite{algarni2008novel}, 
Zernike moment~\cite{wei2014alignment}, and
Gabor and Zernike features~\cite{kong2014novel}.
In contrast, local methods extract discriminative features from local regions of the shoeprint~\cite{alizadeh2021automatic}, making them more adept at handling partial prints.  
For instance, 
\cite{kortylewski2015unsupervised} exploits Wavelet-Fourier transform features, \cite{alizadeh2017automatic} introduces a block sparse representation technique, and \cite{almaadeed2015partial} combines the Harris and the Hessian point of interest detectors with SIFT descriptors. 
Recent works \cite{zhang2017adapting, kong2019cross, ma2019shoe} use features from networks pretrained on ImageNet \cite{deng2009imagenet}.  
However, the lack of large-scale shoeprint datasets hampers their effectiveness. 
To address this, we create a large-scale training dataset by leveraging tread images from online retailers and utilizing an off-the-shelf predictor \cite{shafique2022shoerinsics} to estimate their depth maps and prints.


{\bf Image retrieval.}
Image retrieval techniques have been a popular research problem for several decades~\cite{zheng2017sift}. Traditional methods use handcrafted local features~\cite{lowe2004distinctive, bay2008speeded}, often coupled with approximate nearest-neighbor search methods using KD trees or vocabulary trees~\cite{beis1997shape, han2015matchnet, nister2006scalable, philbin2007object}. 
More recently, the success of CNNs in classification tasks encourages their use in image retrieval tasks~\cite{babenko2014neural, sharif2014cnn}. Global features can be generated by aggregating CNN features~\cite{babenko2015aggregating, tolias2015particular, arandjelovic2016netvlad, gordo2016deep, radenovic2016cnn, tolias2016image, noh2017large, radenovic2018fine, fire}, while local features can also be used for spatial verification~\cite{philbin2007object, noh2017large, cao2020unifying, fire, supcon} which ensure better performance by using geometric information of objects.
Our problem differs from this category of work since our query and database data come from different domains - crime-scene shoeprints and depth maps of shoe treads. Even within our query set of crime-scene shoeprints, images can be from various sources such as blood, dust, and sand impressions. 


{\bf Cross-domain image retrieval.}
More closely related to our work is cross-domain image retrieval (CDIR), where the query and database images come from different domains. 
The fundamental idea is to map both domains into a shared semantic feature space to alleviate the cross-domain gap. 
Learning a distinct representation for each shoe model can be categorized as fine-grained cross-domain image retrieval (FG-CDIR) as we aim to retrieve one instance from a gallery of same-category images. 
It is harder than category-level classification~\cite{dey2019doodle, dutta2019semantically, yelamarthi2018zero} tasks since the differences between shoe treads are often subtle. 
 A  popular problem of this category, fine-grained sketch-based image retrieval (FG-SBIR), was introduced as a deep triplet-ranking based siamese network~\cite{qian2016sketch} for learning a joint sketch-photo manifold. 
FG-SBIR adopts attention-based modules with a higher order retrieval loss~\cite{song2017deep}, textual tags~\cite{chowdhury2023scenetrilogy, song2017fine}, and hybrid cross-domain generation~\cite{pang2017cross}.
The recent work \cite{sketchLVM} leverages a foundation model (CLIP) and 
\cite{zse-sbir} explicitly learns local visual correspondence between sketch and photo to offer explainability. 
These works differ from ours in that we do not have any ground-truth training data from our query domain, and thus have to simulate it as best as we can. Additionally, our aligned query and database images enable us to use spatially-aware techniques like spatial feature masking.





\section{Problem Setup and Evaluation Protocol}

Our goal is to retrieve shoe models that best match crime-scene impressions by comparing against a comprehensive shoe collection. We propose using tread images from online retailers to build our reference database. The problem formulation and evaluation protocol is outlined below.

\begin{figure}[t]
\center
{\small \hspace{.2cm}tread image \hspace{2.3cm} predicted depth \hspace{1.8cm} predicted print \hspace{.4cm}}
  \includegraphics[trim={0.3cm 13.1cm 0.25cm 2cm},clip,width=\linewidth]{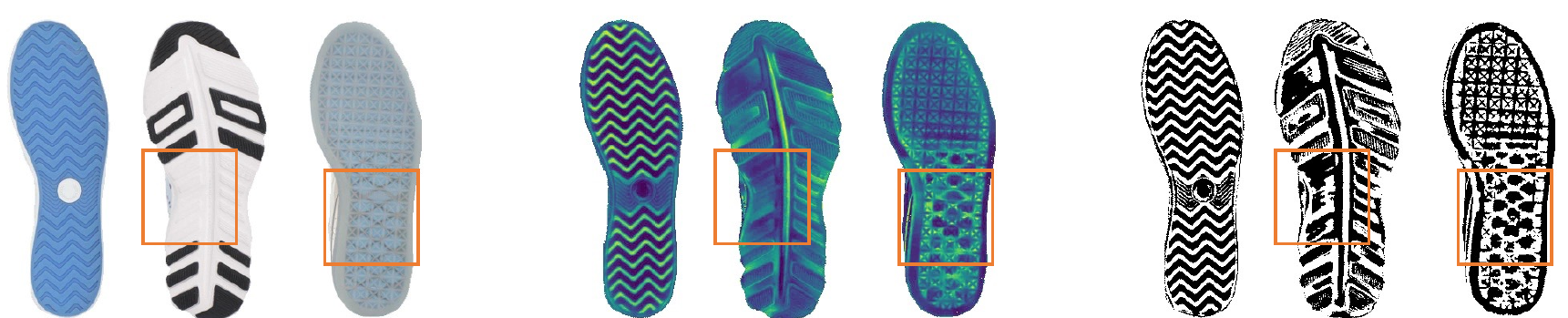}
  \vspace{-6mm}
  \caption{Examples from train-set. 
  We create training data from online retailers and prepare their annotations by predicting their depth maps and prints~\cite{shafique2022shoerinsics},
  although the depth and print predictions are sometimes inaccurate (2nd and 3rd shoe). 
  }
\vspace{-2mm}
\label{fig:training_dataset}
\end{figure}

\subsection{Problem setup}

Given an input \emph{shoeprint} image (\cref{fig:testing_dataset}), our goal is to 
retrieve the most relevant \emph{shoe tread} models from a reference database (\cref{fig:training_dataset}).
{\bf A method should retrieve a ranked list $[r_1, r_2, ..., r_n]$ of shoe models from this database,
where $r_i$ is more likely to leave a crime-scene shoeprint similar to the input shoeprint than $r_j$ for $i<j$}. 
Ranking might involve comparing learned features to represent both shoeprint images and shoe tread examples of the database.
With the retrieved short-list of ranked examples, a crime-scene investigator will then examine them for further judgement.

In our work, we organize the database by storing shoe tread images and their depth maps, as prior work \cite{shafique2022shoerinsics} demonstrates that using depth allows synthesizing shoeprint images as training data (cf. \cref{subsection:train-set}).
Hence, we create such a database.
Consequently, methods should 
(1) address the domain gap between crime-scene shoeprints and clean shoe tread depth maps, and 
(2) match partly visible shoeprints to corresponding regions of shoe tread depth maps.


\subsection{Evaluation Protocol}
\label{subsec:evaluation_protocol}

To benchmark methods, we introduce two validation sets of crime-scene shoeprints with ground-truth shoe model labels, which are linked to a large-scale reference database (see details in \cref{subsection:validation_set}). 
Note that the ground-truth for a shoeprint may contain multiple shoe models since tread patterns can be shared by different shoe models. 
In practice, we expect a human-in-the-loop approach: crime-scene investigators will look through the top $K$ retrieved shoe models.
Such a practice will greatly mitigate an open-set issue, i.e., 
finding that an input shoeprint does not have similar shoe models in the current database.
We set $K$ to be a realistically small value of 100, representing the top $0.4\%$ shoe models in our reference database. 
We use two metrics to compare models based on their top $K$ retrievals.
Our first metric, mean average precision at K (mAP@K), is a standard metric to compare ranking performance. It considers both the number of positive matches and their positions in the ranking list. 
The second metric, hit ratio at K (hit@K), is more intuitive and represents the fraction of times we get at least one positive match in the top $K$ retrievals. This metric is useful because a positive match can be used in a query expansion step to retrieve other good matches much more effectively~\cite{efthimiadis1996query}. 
Both metrics have values between 0 and 1, with higher numbers representing better performance. 
The supplement has further details.

\begin{figure}[t]
\centering
  \includegraphics[trim={0 14.5cm 8.7cm 2.3cm},clip,width=\linewidth]{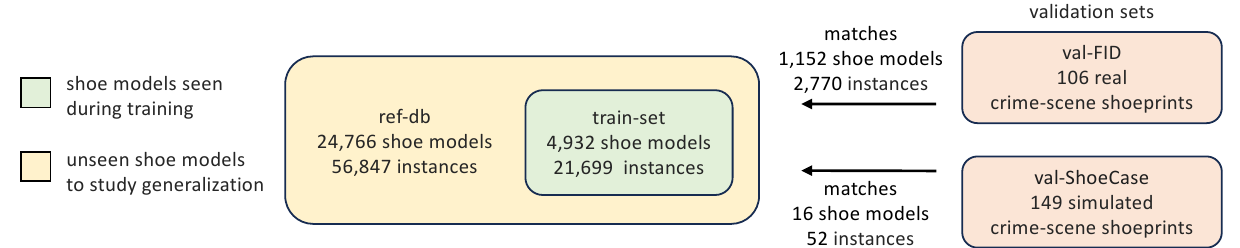}
  \vspace{-6mm}
  \caption{Dataset statistics. We have a reference database (ref-db) and two validation sets (val-FID and val-ShoeCase) with crime-scene impressions to query against ref-db. 
  We use a section of ref-db for training (train-set) and leave the rest to study generalization. 
  Ground-truth labels from our validation sets connect our query crime-scene shoeprints to shoes in ref-db. See details in \cref{sec:datasets} and visual examples in \cref{fig:training_dataset} and~\ref{fig:testing_dataset}.
  }
\label{fig:dataset_statistics}
\end{figure}

\section{Dataset Preparation}
\label{sec:datasets}

\begin{figure}[t]
{\footnotesize \raggedright \hspace{1.1cm} FID-crime  \hspace{1.8cm} FID-clean \hspace{0.9cm} ShoeCase-blood  \hspace{0.2cm} ShoeCase-dust\\}
\vspace{0.05cm}
  \includegraphics[trim={.1cm 13cm 0.1cm 2.4cm},clip,width=\linewidth]{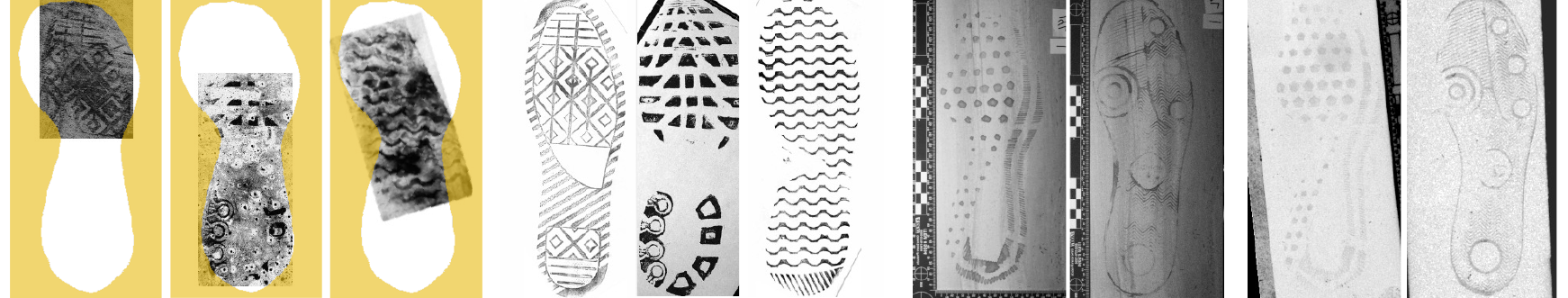}
  \vspace{-6mm}
  \caption{Examples from val-FID and val-ShoeCase.
  Val-FID contains real crime-scene prints (FID-crime) and clean, fully visible lab impressions (FID-clean). We show FID-crime and FID-clean shoeprints corresponding to the same shoe models for easier comparison. Note that we show a yellow shoe outline on the FID-crime prints for visualization purposes and the outline does not exist in FID-crime images.
  Val-ShoeCase contains simulated crime-scene shoeprints on blood (ShoeCase-blood) and dust (ShoeCase-dust). All val-ShoeCase prints are full-sized, as opposed to val-FID.  
  }
\vspace{-3mm}
\label{fig:testing_dataset}
\end{figure}

We train our model on a dataset (train-set) of aligned shoe tread depth maps and clean shoeprints. 
To study the effectiveness of models, we introduce a large-scale reference database (ref-db) of tread depth maps, along with two validation sets (val-FID and val-ShoeCase) created by reprocessing existing datasets of crime-scene shoeprints~\cite{fid300, tibben2023shoecase}. We match shoeprints from the validation sets to ref-db and add labels connecting shoeprints in val-FID and val-ShoeCase to ref-db to enable quantitative analysis.
An overview of the datasets is provided in \cref{fig:dataset_statistics}, while \cref{fig:training_dataset} and \cref{fig:testing_dataset} present example depth maps, clean prints, and crime-scene prints. In this section, we elaborate on our training dataset (train-set), reference database (ref-db), and validation sets (val-FID and val-ShoeCase).

\subsection{Online Shoe Tread Depth Maps and Prints for Training}
\label{subsection:train-set}

{\bf Train-set.}
Online retailers~\cite{Sixpm, Zappos} showcase images of shoe treads for advertisement. Our training set (train-set) contains depth maps and clean, fully visible prints from such tread images as predicted by~\cite{shafique2022shoerinsics}. We also apply segmentation masks as suggested by~\cite{shafique2022shoerinsics} to the predictions.
To ensure consistency across all images, we employ a global alignment method to minimize variations in scale, orientation, and center using a simple model. 
\cref{fig:training_dataset} displays some sample shoe-tread images along with their corresponding depth and print predictions.
Online retailers categorize shoe styles using stock keeping units (SKUs), which we use as shoe model labels. Shoes with the same SKUs can have different colors and sizes.
Different shoe models may share the similar tread pattern, making them appear to be duplicates; we do not remove such likely duplicates as investigators will still examine them from the retrieved examples for the final judgement.

{\bf Statistics.}
Train-set contains 21,699 shoe instances from 4,932 different shoe models. 
Each shoe model in our database can have shoe-tread images from multiple shoe instances, possibly with variations in size, color, and lighting. The tread images in train-set have a resolution of 384$\times$192. 

{\bf Inaccuracies.} It is important to note that the training dataset can have some inaccuracies since it comes from raw data downloaded from online retailers. Some tread images might have incorrect model labels, and some images may not depict shoe treads. Other inaccuracies come from imperfect depth and print prediction (cf. \cref{fig:training_dataset}), segmentation errors, and alignment failures. 
We hope to mitigate the errors by including multiple instances per shoe model in train-set. 

\subsection{Reference Database and Crime-scene Shoeprints for Validation}
\label{subsection:validation_set}

{\bf Ref-db.} 
We introduce a reference database (ref-db) by extending train-set to include more shoe models. The added shoe models are used to study generalization to unseen shoe models. Ref-db contains a total of 56,847 shoe instances from 24,766 different shoe models. 
The inclusion of multiple instances per shoe model in ref-db allows the depth predictor some margin for error  (cf. \cref{fig:training_dataset}), ensuring minimal impact on the overall matching algorithm performance
since it has multiple chances to match a query print to a shoe model. 
The supplement has details on the distribution of shoe models from our validation sets in ref-db.

{\bf Val-FID.}
We reprocess the widely used FID300~\cite{fid300} to create our primary validation set (val-FID). 
Val-FID contains real crime-scene shoeprints (FID-crime) and a corresponding set of clean, fully visible lab impressions (FID-clean). Examples of these prints are shown in \cref{fig:testing_dataset}.
The FID-crime prints are noisy and often only partially visible. It contains impressions made by blood, dust, etc on various kinds of surfaces including hard floors and soft sand. 
To ensure alignment with ref-db, we preprocess FID-crime prints by placing the partial prints in the appropriate position on a shoe ``outline'' (cf. \cref{fig:testing_dataset}), a common practice in shoeprint matching during crime investigations.

We manually found matches to 41 FID-clean prints in ref-db by visual inspection. These are all unique tread patterns and correspond to 106 FID-crime prints.  
Given that multiple shoe models in ref-db can share the same tread pattern, we store a list of target labels for each shoeprint in FID-crime.
These labels correspond to 1,152 shoe models and 2,770 shoe instances in ref-db (cf. 
\cref{fig:dataset_statistics}).

{\bf Val-ShoeCase.}
We introduce a second validation set (val-ShoeCase) by reprocessing ShoeCase~\cite{tibben2023shoecase} which consists of simulated crime-scene shoeprints made by blood (ShoeCase-blood) or dust (ShoeCase-dust) as shown in \cref{fig:testing_dataset}.
These impressions are created by stepping on blood spatter or graphite powder and then walking on the floor. The prints in this dataset are full-sized, and we manually align them to match ref-db. 


ShoeCase uses two shoe models (Adidas Seeley and Nike Zoom Winflow 4), both of which are included in ref-db. The ground-truth labels we prepare for val-ShoeCase include all shoe models in ref-db with visually similar tread patterns as these two shoe models since we do not penalize models for retrieving
shoes with matching tread patterns but different shoe models.
Val-ShoeCase labels correspond to 16 shoe models and 52 shoe instances in ref-db (cf. \cref{fig:dataset_statistics}). 



\section{Methodology}
\label{sec:methodology}

\begin{figure*}[t]
{\small \raggedright
\hspace{3.5mm} occlusion  \hspace{6.5mm} erasure \hspace{9.6mm} noise \hspace{10.5mm} simulated crime-scene shoeprints $\hat S$}
\vspace{-3mm}
\center
  \includegraphics[trim={0.2cm 15.2cm 7.9cm 2.2cm},clip,width=\linewidth]{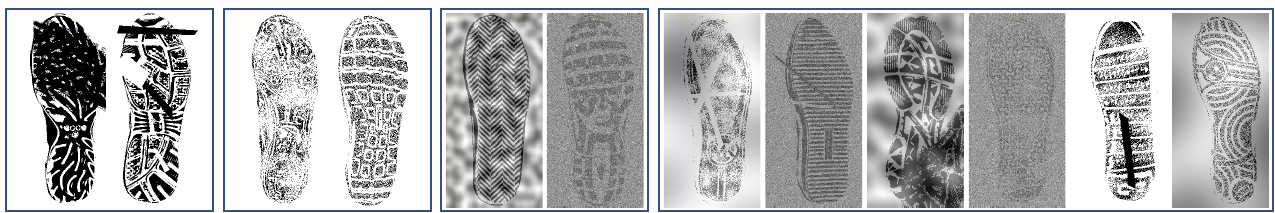}
  \vspace{-7mm}
  \caption{Examples of data augmentation. Our data augmentation module $Aug$ simulates crime-scene shoeprints (cf. \cref{fig:testing_dataset}) from clean, fully visible prints in our training set (cf. \cref{fig:training_dataset}). 
  $Aug$ optionally (1) introduces occlusion such as overlapping prints and random shapes, (2) erases parts of the print to create a grainy appearance, and (3) adds noise to mimic background clutter. 
  }
\label{fig:noise_augmentation_3}
\vspace{-5mm}
\end{figure*}

In this section, we introduce \emph{CriSp}, our representation learning framework to match crime-scene shoeprint images $S$ to tread depth maps $d$. 
An overview of our training pipeline is shown in \cref{fig:training_overview}. 
\emph{CriSp} is trained using a dataset of globally aligned
tread depth maps $d$ and clean, fully-visible shoeprints $s$  (see details in \cref{subsection:train-set}). 
The main components of our pipeline are
(1) a data augmentation module $Aug$ that simulates crime-scene shoeprints,
(2) an encoder network $Enc$ that maps depths and shoeprints to a spatial feature representation, and (3)
a spatial masking module $M$ that masks out irrelevant portions from partially visible shoeprints.


{\bf Data augmentation.}
Our data augmentation module $Aug$ simulates noisy and occluded crime-scene shoeprints (cf. \cref{fig:testing_dataset}) from clean, fully-visible prints (cf. \cref{fig:training_dataset}), denoted as $\hat S = Aug(s)$. 
$Aug$ uses three kinds of degradations (occlusion, erasure, and noise) as visualized in \cref{fig:noise_augmentation_3}. Occlusion can be in the form of overlapping prints or random shapes. 
Erasures achieve the grainy texture of crime-scene prints and noise adds background clutter to the images. 
Further details are provided in the supplement. 

{\bf Encoder for spatial features.}
Our encoder $Enc$ maps tread depths $d$ and simulated crime-scene shoeprints $\hat S$ to a feature representation $z$, denoted as $z=Enc(x)$ where $x \in [d, \hat S]$. $Enc$ consists of a modified ResNet50~\cite{he2016deep} with the final pooling and flattening operation removed followed by a couple of convolution layers. 
$Enc$ produces features of shape $[C, H, W]$ where $C$ is the feature length ($C=128$ in our work), and $H$ and $W$ are the encoded height and width, respectively. 
As our training data and query prints are globally aligned (cf. \cref{sec:datasets}), $Enc$ allows access to features at each (course) spatial location of the image, facilitating comparisons in corresponding locations of shoe treads. 
$Enc$ has two input channels for depth and print, respectively. It processes only one input at a time and pads the other input channel with zeros. 

{\bf Spatial feature masking.}
During training, we simulate partially visible crime-scene shoeprints by applying a random rectangular mask $m$ to query prints. 
Our feature masking module $M$ applies a corresponding mask to spatial features $z$ to obtain $\bar z = M(z, m)$.
$M$ resizes mask $m$ to a dimension of $[H, W]$, uses it to zero out spatial features outside the mask, and normalizes the masked features.
This allows our model to focus on the visible portion of the prints.
While it would make sense to apply mask $m$ to tread depth images as well, we opt not to do this as it would necessitate recomputing all the database depth features for each query print image at inference time, which is not scalable.

{\bf Training loss and similarity metric.}
We train our model using supervised contrastive learning~\cite{supcon}, which extends self-supervised contrastive learning to a fully supervised setting to learn from data using labels. 
For a set of $N$  depth/print pairs $\{d_k, s_k\}_{k=1 \dots N}$
from shoe models $\{l_k\}_{k=1 \dots N}$ within a batch, and a randomly generated mask $m$ per batch, we compute masked spatial features $\{\bar z_i\}_{i=1 \dots 2N}$ and corresponding shoe labels $\{\bar l_i\}_{i=1 \dots 2N}$ where 
$\bar z_{2k}=M(Enc(d_k), m)$, $\bar z_{2k+1}=M(Enc(Aug(s_k)), m)$, and $\bar l_{2k} = \bar l_{2k+1} = l_{k}$. We treat $\bar z$ as a vector of size $CHW$ and apply the following loss.
\begin{align}\small
\mathcal{L} = \sum\limits_{i \in I } \frac{-1}{|P(i)|}  \sum\limits_{p \in P(i) } \text{log} \frac{\text{exp}(\bar z_i \cdot \bar z_p / \tau)}{\sum\limits_{a \in A(i)} \text{exp}(\bar z_i \cdot \bar z_a / \tau)}
\end{align}
Here,  $i \in I \equiv \{1 \dots 2N\}$, 
$A(i) \equiv I \setminus \{i\}$, and 
$P(i) \equiv \{p \in A(i) : \bar l_p = \bar l_i\}$ is the set of indices of all positives in the batch distinct from $i$.
$|P(i)|$ is the cardinality of $P(i)$.  
The $\cdot$ symbol denotes the inner product, and 
$\tau \in \mathcal{R}^+$ is a scalar temperature parameter.
This loss corresponds to using cosine similarity to measure similarity between images.

{\bf Sampling.}
For the above loss to be effective, we must have (enough) positive examples  within a batch.
However, if we uniformly sample shoe models from the large-scale dataset of a large number of shoe models, 
a training batch might contain unique shoe models that does not have pairs of positive examples.
Therefore, 
we sample training data in pairs, 
i.e. we choose $N / 2$ shoe models randomly and select two random instances from each shoe model.

\section{Experiments}
\label{sec:experiments}

We evaluate our \emph{CriSp} and compare it with state-of-the-art methods on automated shoeprint matching \cite{kong2019cross} and image retrieval~\cite{supcon, fire, sketchLVM, zse-sbir}.
We begin with 
visual comparison and quantitative evaluation, followed by an ablation study and analysis of our design choices.
We release our dataset and make our code publicly available at  \url{https://github.com/Samia067/CriSp}.


\begin{figure*}[t!]
\center
  \ query \hspace{35mm} \emph{CriSp} top-10 retrievals \hspace{37mm} \ \ \\
  \includegraphics[trim={0cm 2.7cm 14.5cm 5.25cm},clip,width=\linewidth]{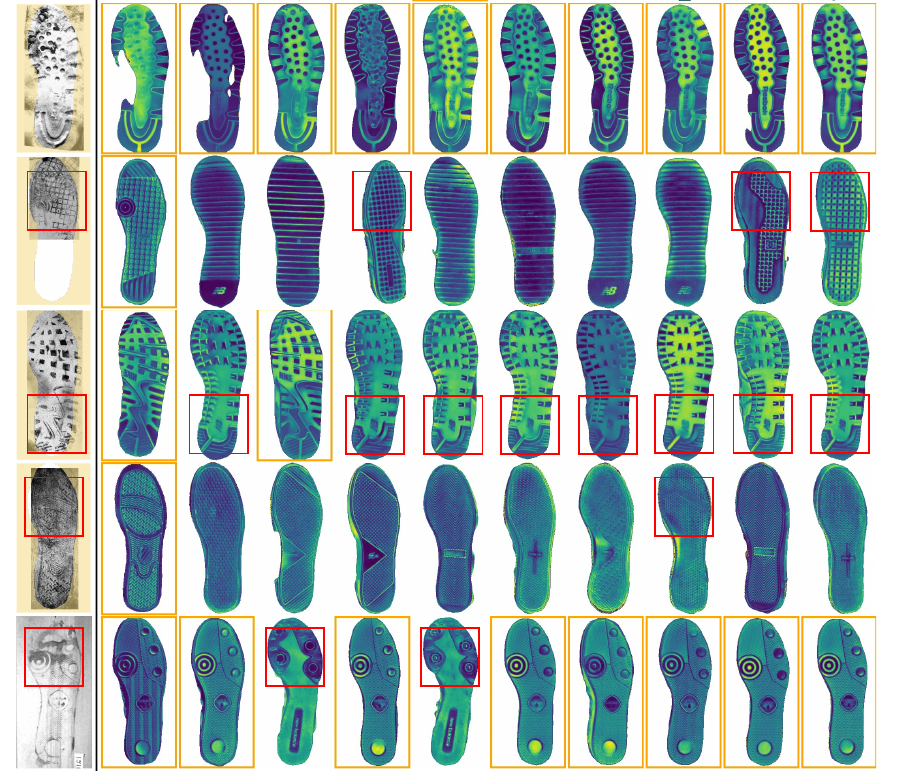}
  \vspace{-7.5mm}
  \caption{
  Visualization of the top 10 retrievals by \emph{CriSp} on val-FID (rows 1-4) and val-ShoeCase (row 5). 
  \emph{CriSp} retrieves positive matches (highlighted by orange frames) even when crime-scene shoeprints have very limited visibility or severe degradation. 
  Additionally, corresponding locations on the retrieved shoes share similar patterns to the query print, even in negative matches (marked by red boxes). 
  }
\label{fig:qual_results}
\vspace{-5mm}
\end{figure*}

\begin{figure*}[t!]
\centering
  \ query \hspace{8mm} \emph{CriSp}   \hspace{14mm} ZSE-SBIR  \hspace{13mm} SketchLVM \hspace{12mm}  FIRe \hspace{8mm} \ \\
  \includegraphics[trim={0cm 3.8cm 11.55cm 6.86cm},clip,width=\linewidth]{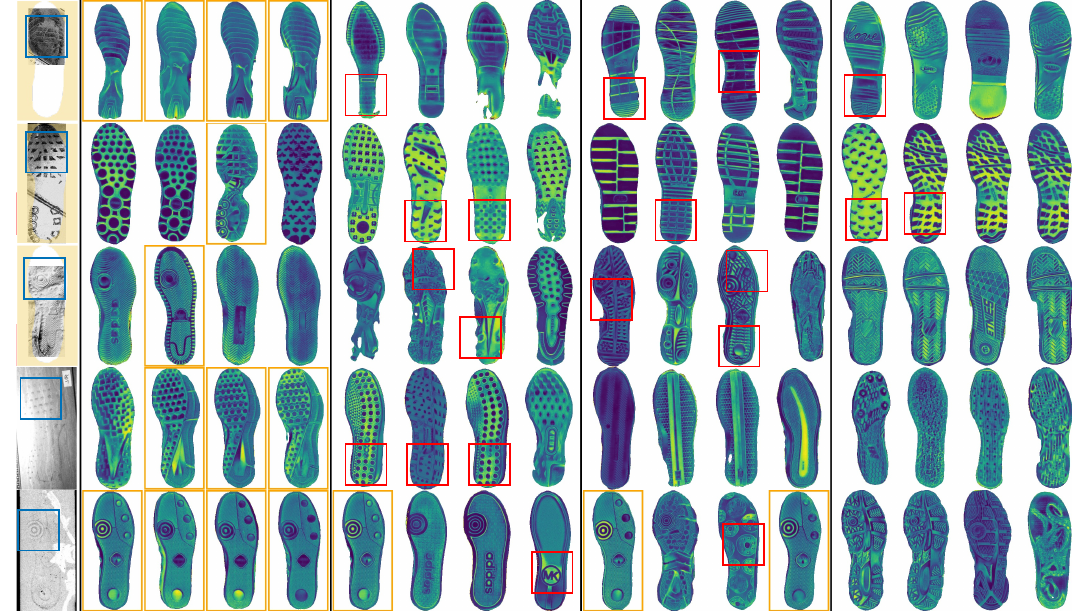}
  \vspace{-6mm}
  \caption{
  Qualitative comparison with state-of-the-art methods on val-FID (rows 1-3), val-ShoeCase (rows 4-5).
  We show the top 4 retrieved results. \emph{CriSp} demonstrates the ability to localize patterns, allowing it to achieve more precise retrievals (highlighted by orange frames) than previous methods.
  While prior methods identify similar patterns to the query print (cf. blue regions on query images), they cannot determine if they are from corresponding locations, as indicated by the red boxes in retrieved images.
  }
\label{fig:qual_comparison}
\vspace{-2mm}
\end{figure*}

\subsection{Qualitative Results of CriSp}

\cref{fig:qual_results} shows the top 10 retrievals of our method \emph{CriSp} on the val-FID and val-ShoeCase datasets. Notable, \emph{CriSp} can retrieve a positive match very early even when the shoeprint has significantly limited visibility or is severely degraded. These retrievals show how \emph{CriSp} effectively matches distinctive patterns from corresponding regions of the tread. 
\cref{fig:qual_comparison} shows a comparison with related methods fine-tuned on our dataset. Clearly, \emph{CriSp} performs significantly better at retrieving positive matches early.
See more visualizations in the supplement.

\subsection{Comparison with State-of-the-art}

\emph{CriSp} consistently outperforms previous methods across most validation examples (details in the supplement). 
Table~\ref{table:comparison_to_state_of_the_art_fid} and \ref{table:comparison_to_state_of_the_art_shoecase} list comparisons on our two evaluation metrics introduced in \cref{subsec:evaluation_protocol}. We analyze these results below.

{\bf Comparison with shoeprint matching.}
MCNCC~\cite{kong2019cross} employs features from pretrained networks on ImageNet for automated shoeprint matching. However, leveraging learning on shoeprint-specific data, \emph{CriSp} exhibits superior performance on both val-FID (see \cref{table:comparison_to_state_of_the_art_fid}) and val-ShoeCase (see \cref{table:comparison_to_state_of_the_art_shoecase}). Although MCNCC proposes to use clean shoeprint impressions as the reference database to match with, we use tread depth maps to be consistent with other methods and to achieve enhanced results. More details are in the supplement.  

\begin{table}[t!]
\small
\caption{
{\bf Benchmarking results on real crime-scene prints from val-FID}. 
We use hit@100 and mAP@100 as the metrics and compare previous methods trained on our dataset with / without data augmentation (cf. \cref{sec:methodology}).
Recall that our proposed data augmentation simulates crime-scene shoeprints from clean, fully-visible prints for the training examples. 
Clearly, all other prior methods benefit greatly from using our data augmentation technique. 
MCNCC achieves low mAP because it (1) uses off-the-shelf features from an ImageNet-pretrained model, which is not tailored to shoeprint matching,
and (2) works on a more challenging and larger database (56,847 images) in our work, compared to the small-scale one (1,175 images) in its original paper~\cite{kong2019cross}.
SupCon also performs poorly as it samples data uniformly from the large training set that fails to guarantee enough positive pairs in training batches. Our modification (which is row-1 in \cref{table:feature_masking}) ensures enough positive pairs in batches through careful data sampling, yielding significant improvements.
Lastly, \emph{CriSp} significantly outperforms all the compared methods. 
}
\vspace{-3mm}
\setlength\tabcolsep{0pt} 
\begin{tabular*}{\linewidth}{@{\extracolsep{\fill}}*{1}{r}@{\extracolsep{2mm}}*{1}{l}@{\extracolsep{\fill}}*{4}{c}} 
\toprule
\multicolumn{2}{c}{\multirow{2}{*}{method}} & \multicolumn{2}{c}{w/o our data aug} & \multicolumn{2}{c}{w/ our data aug} \\
 & & hit@100 & mAP@100 & hit@100 & mAP@100\\
\midrule
         IJCV'19 & MCNCC~\cite{kong2019cross} &  0.0849 & 0.0018 & - & - \\   
         NeurIPS'20 & SupCon~\cite{supcon} &  0.0472 & 0.0020 & 0.0755 & 0.0096 \\  
         ICLR'21 & FIRe~\cite{fire} &  0.1132 &  0.0014 &  0.2075 &  0.0398\\       
         CVPR'23 & SketchLVM~\cite{sketchLVM} &  0.0849&  0.0066 &  0.1981 & 0.0384 \\ 
         CVPR'23 & ZSE-SBIR~\cite{zse-sbir} &  0.0943 & 0.0065 & 0.4528 & 0.1412  \\  
         & \textbf{CriSp} &  0.0754 & 0.0174 & \textbf{0.5472} & \textbf{0.2071}\\  
\bottomrule
\end{tabular*}
\vspace{-2mm}
\label{table:comparison_to_state_of_the_art_fid}
\end{table}

\begin{table}[t!]
\small
\caption{
{\bf Benchmarking results on simulated crime-scene prints from val-ShoeCase}, which includes shoeprints made by blood and dust. We use hit@100 and mAP@100 as the metrics.
\emph{CriSp} performs the best across print categories.  
All prior methods have been fine-tuned on our dataset using our data augmentation technique, as they perform poorly otherwise (cf. \cref{table:comparison_to_state_of_the_art_fid}). 
Note that both ZSE-SBIR and CriSp coincidentally achieve positive matches on 62 blood prints (62/77 = 0.8052) and 68 dust prints (68/72 = 0.9444), resulting in the same hit@100, which measures the fraction of times a method gets at least one positive match within the top 100 retrievals. 
}
\vspace{-3mm}
\setlength\tabcolsep{0pt} 
\setlength{\tabcolsep}{.267em} 
\begin{tabular*}{\linewidth}{@{\extracolsep{\fill}}*{1}{l}@{\extracolsep{\fill}}*{4}{c}}
\toprule
\multirow{2}{*}{method}  & \multicolumn{2}{c}{ShoeCase-blood} & \multicolumn{2}{c}{ShoeCase-dust} \\
 & hit@100 & mAP@100 & hit@100 & mAP@100 \\
\midrule
         MCNCC~\cite{kong2019cross} & 0.0000 & 0.0000 &  0.0000 & 0.0000 \\   
         SupCon~\cite{supcon}       & 0.0000             & 0.0000 &  0.0000 & 0.0000  \\
         FIRe~\cite{fire}           &  0.3896            & 0.0275 & 0.8194 & 0.3779 \\       
         SketchLVM~\cite{sketchLVM} & 0.6623             & 0.1058 & 0.5972 & 0.2696  \\ 
         ZSE-SBIR~\cite{zse-sbir}   & \textbf{0.8052}    & 0.1849 & \textbf{0.9444} & 0.4063 \\ 
         \textbf{CriSp}              &  \textbf{0.8052}   & \textbf{0.4355} & \textbf{0.9444} & {\bf 0.6792} \\   
\bottomrule
\end{tabular*}
\vspace{-5mm}
\label{table:comparison_to_state_of_the_art_shoecase}
\end{table}

{\bf Comparison with image retrieval.}
\Cref{table:comparison_to_state_of_the_art_fid} and \ref{table:comparison_to_state_of_the_art_shoecase} demonstrate how our \emph{CriSp} consistently outperforms state-of-the-art methods in image retrieval (SupCon~\cite{supcon}, FIRe~\cite{fire}, SketchLVM~\cite{sketchLVM}, ZSE-SBIR~\cite{zse-sbir}).
We fine-tune these methods on our training data containing tread depth maps and clean, fully-visible shoeprints. Additionally, we use our data augmentation module $Aug$ to simulate crime-scene shoeprints while training prior methods as the wide domain gap between crime-scene prints and the training data causes them to perform poorly otherwise (\cref{table:comparison_to_state_of_the_art_fid}). 
Even when prior methods use our data augmentation, \emph{CriSp} significantly outperforms them on both val-FID (\cref{table:comparison_to_state_of_the_art_fid}) and val-ShoeCase (\cref{table:comparison_to_state_of_the_art_shoecase}).
The ablation study (\cref{table:feature_masking}) shows that our spatial feature masking technique greatly improves the performance. 
Qualitative comparison on both validation sets in \cref{fig:qual_comparison}
also confirm that \emph{CriSp} is better able to match shoeprint patterns to corresponding locations on tread depth maps, thus making positive retrievals early. This is reflected by our mAP@100 values when compared to prior methods on both validation sets (\cref{table:comparison_to_state_of_the_art_fid} and \ref{table:comparison_to_state_of_the_art_shoecase}).

{\bf Scalability.}
In practice, when dealing with a large reference database, scalability becomes crucial. 
Unlike our closest competitor ZSE-SBIR~\cite{zse-sbir}, which necessitates the recomputation of all database features for each query, \emph{CriSp} offers a scalable solution.
It can precompute spatial database features and efficiently perform feature masking and cosine similarity calculations for each query, enabling rapid retrieval even with extensive reference databases.

{\bf Simulating partial print.}
Retrievals by prior methods on partial shoeprints in \cref{fig:qual_comparison}
reveal instances of poorly segmented tread depth maps, where significant portions of the tread pattern have been erased. 
This raises the question of whether prior methods would exhibit improved performance if trained with masks simulating partial prints.
However, it is worth noting that prior methods perform better when trained without such masks, as detailed in the supplement.

{\bf Val-FID versus val-ShoeCase.}
Methods show a wider variation in performance on Val-ShoeCase than val-FID. This discrepancy arises from the fact that val-FID contains the diversity of real crime-scene shoeprints, while val-ShoeCase systematically simulates crime-scene prints.
Additionally, val-ShoeCase contains prints from shoe models with only two unique tread patterns while val-FID contains prints from 41 unique tread patterns (cf. \cref{subsection:validation_set}).

\begin{table}[t]
\parbox{.45\linewidth}{
\caption{
{\bf Testing database image configurations}. 
The hit@100 and mAP@100 values for FID-clean shoeprints indicate that using only tread depth as the database image configuration yields the best performance. 
Results for FID-crime are not reported in this experiment as we do not simulate crime-scene prints.
}
\vspace{-3mm}
\setlength\tabcolsep{0pt} 
\begin{tabular*}{\linewidth}{@{\extracolsep{\fill}}*{5}{c}} 
\toprule
\multicolumn{3}{c}{Database config.} & \multicolumn{2}{c}{FID-clean} \\ 
\cmidrule{1-3}
\cmidrule{4-5}
RGB & depth & print & hit@100 & mAP@100  \\
\midrule
\checkmark & & & 0.195 & 0.066\\
& \checkmark & & \textbf{0.512} & \textbf{0.203} \\
& & \checkmark & 0.171 & 0.015 \\
\checkmark & \checkmark & \checkmark & 0.293 & 0.057\\
\bottomrule
\end{tabular*}
\label{table:database_input_types}
}
\hfill
\parbox{.5\linewidth}{
\caption{{\bf Ablation of data augmentation techniques}. 
We train ResNet50 networks using techniques of our data augmentation and report hit@100 and mAP@100 on FID-crime shoerpints.
Results confirm that each technique (visualized in \cref{fig:noise_augmentation_3}) individually improves retrieval results and performs best when used together.
}
\vspace{-3mm}
\setlength\tabcolsep{0pt} 
\begin{tabular*}{\linewidth}{@{\extracolsep{\fill}}*{5}{c}} 
\toprule
\multicolumn{3}{c}{Data augmentation} & \multicolumn{2}{c}{FID-crime} \\ 
\cmidrule{1-3}
\cmidrule{4-5}
occlusion & erasure & noise & hit@100 & mAP@100 \\
\midrule
 & & &   0.009 &	0.0000 \\
\checkmark & &  & 0.019 & 0.0003 \\
& \checkmark & & 0.075 & 0.0098\\
& &  \checkmark & 0.170 & 0.0241\\
\checkmark & \checkmark & \checkmark & \textbf{0.226} & \textbf{0.0520} \\
\bottomrule
\end{tabular*}
\label{table:data_augmentation}
}
\vspace{-5mm}
\end{table}

\begin{table}[t]
    \centering
    \caption{
    {\bf Ablation of spatial features and feature masking}.
    We validate the effect of using spatial features and applying feature masking on either our encoder $Enc$, which incorporates spatial features during training, or a pretrained ResNet50 which is trained with our data augmentation (cf. \cref{table:data_augmentation}).
    With ResNet50 that does not utilize spatial features during training, we obtain spatial features by removing the last pooling operation.  
    We report  hit@100 and mAP@100 metrics for FID-crime shoeprints from val-FID using.
    Using spatial features from a pretrained ResNet50 boosts retrieval performance. 
    Moreover, masking the spatial features improves performance further for both the ResNet50 and our $Enc$. 
    Lastly, adding query print masking during training performs the best, yielding hit@100=0.5472 and mAP@100=0.2071.
    }
\vspace{-3mm}
    \setlength\tabcolsep{0pt}
    \begin{tabular*}{\linewidth}{@{\extracolsep{\fill}}*{7}{c}} 
  \toprule
  \multirow{2}{*}{encoder} & \multirow{2}{*}{\parbox{2cm}{\centering  train w/ spatial feat.}} & \multirow{2}{*}{\parbox{1.5cm}{\centering  spatial features}} & \multirow{2}{*}{\parbox{1.5cm}{\centering mask features}} & \multirow{2}{*}{\parbox{2cm}{\centering  mask query print}} & \multicolumn{2}{c}{FID-crime} \\ 
  \cmidrule{6-7}
         & &  &   & &  hit@100 & mAP@100 \\ 
        \midrule
        ResNet50 & & &  & &  0.2264 & 0.0520 \\ 
        ResNet50 & & \checkmark & & &  0.3585 & 0.0863 \\ 
        ResNet50 & & \checkmark & \checkmark & & 0.4245 & 0.1212 \\
        \midrule
        $Enc$ & \checkmark & \checkmark & &  & 0.3774 & 0.1137 \\ 
        $Enc$ & \checkmark & \checkmark & \checkmark &  & 0.4528 & 0.1765 \\
        $Enc$ & \checkmark & \checkmark & \checkmark & \checkmark & \textbf{0.5472}  &  \textbf{0.2071} \\ 
    \bottomrule
    \end{tabular*}
    \label{table:feature_masking}
\vspace{-5mm}
\end{table}

\subsection{Design Choices and Ablation Study}
\label{subsection:design_choices}

We conduct a study of our design choices by training a ResNet50 with a supervised contrastive loss and then sequentially adding modules to investigate their performance impact. Specifically, we analyze database image configurations, data augmentation techniques, and spatial feature masking.

{\bf Database image configuration.}
\label{subsec:exp_input_type_compare}
We start by testing the effectiveness of different types of database image configurations (RGB tread images, depth, and print).
Our analysis shows that depth is the most relevant and informative modality, yielding the best results when used alone (\cref{table:database_input_types}). 
Print can be derived from depth by  thresholding~\cite{shafique2022shoerinsics} and the extra information in rgb tread images (lighting and albedo) can be  distracting. 

{\bf Data augmentation.}
Next, we test the effectiveness of each component of our data augmentation technique. \Cref{table:data_augmentation} shows that all 3 components contribute to improved performance and work best when used together, bringing our hit@100 and mAP@100 on FID-crime to (0.226, 0.0520) from (0.009, 0.000).

{\bf Spatial features and feature masking.}
With our data augmentation in place, we study the effect of spatial feature masking, which helps \emph{CriSp} match query print patterns to the relevant spatial locations of the database tread depth maps.
Table~\ref{table:feature_masking} shows the influence of using spatial features and feature masking. 
Our findings indicate that spatial features, feature masking, and query image masking during training all contribute greatly to improving performance.


\section{Discussions and Conclusions}
\label{sec:conclusion}


{\bf Ethics and societal impacts.} 
Our work is motivated by the larger goal of understanding the informational value that shoe tread pattern evidence provides in criminal investigations and forensic examination. We believe that a large dataset of tread patterns and retrieval methods will provide a positive impact as a useful resource for further studies on the human factors and uncertainty involved in making footwear-match likelihood determinations. 

Court systems and footwear examiners do not generally consider matching of shoe make and model as personally identifying information (many people own the same brand of shoe) and rely on further detailed examination of acquired characteristics in conjunction with other evidence to limit false-positives. Nevertheless, there are serious broader concerns about the perils of applying artificial intelligence-based tools in the criminal justice system~\cite{malek2022criminal}. Similar to image retrieval in other domains, we have shown high accuracy in matching shoe tread patterns to query crime-scene evidence, but our research does not address challenging trade-offs that exist between accuracy and fairness in criminal justice risk assessments~\cite{berk2021fairness}.

We thus believe that directly applying automated shoe print retrieval methods in the real world without rigorous justification raises critical ethical issues. Ameliorating such risks in the criminal justice domain requires joint efforts from multiple communities including artificial intelligence, forensic science, criminal justice, legislative science, etc.~\cite{zavrvsnik2020criminal}. We hope our work solicits more attention from these communities and helps foster careful application of AI-based tools (e.g., shoe print matching techniques developed in our work).



{\bf Limitations.}
While \emph{CriSp} significantly outperforms prior methods on this problem, it still has some limitations.
We use CNNs in our work as it is straightforward to apply the proposed spatial feature masking,
yet transformer networks might perform better but it is non-trivial to mask out spatial regions in feature maps.
Our work assumes that the crime-scene shoeprints are manually aligned ahead of time; methods that do not require this might be desired in the future.

{\bf Conclusion.}
In this paper, we propose a method to retrieve and rank the closest matches to crime-scene shoeprints from a database of shoe tread images. This is a socially important problem and helps forensic investigations. 
We introduce a way to learn from large-scale data and propose a spatial feature masking method to localize the search for patterns over the shoe tread. 
Our method consistently outperforms the state-of-the-art on both image retrieval and crime-scene shoeprint matching methods on our two validation sets that we reprocess from the widely used FID and more recent ShoeCase datasets. 

{\bf Acknowledgements.}
This work was funded by the Center for Statistics and Applications in Forensic Evidence (CSAFE) through Cooperative Agreements, 70NANB15H176 and 70NANB20H019.
Shu Kong is partially supported by the University of Macau (SRG2023-00044-FST).




%
%
\bibliographystyle{splncs04}
\bibliography{egbib}

\clearpage


\begin{center}
{\Large\bf{Appendix}\\
}
\end{center}

\vspace{20pt}

\section{Outline}

Our aim is to identify shoe models resembling crime-scene impressions by comparing them to a comprehensive shoe database. Leveraging tread images from online retailers, we construct our reference database, prioritizing tread depth maps over RGB tread images for their greater relevance and informativeness \cite{shafique2022shoerinsics}.
As there is no dataset of crime-scene shoeprints paired with ground-truth tread depth maps, we propose learning from tread depth maps and clean shoeprints predicted from RGB tread images instead. 
We utilize a data augmentation module $Aug$ to bridge the domain gap between clean and crime-scene prints, and a spatial feature masking strategy (using spatial encoder $Enc$ and masking module $M$) to match shoeprint patterns with corresponding locations on tread depth maps. \emph{CriSp} achieves significantly better retrieval results than prior methods.

In this supplementary document, we discuss the following topics:
\begin{itemize}
    \item \Cref{sec:qual_analysis_and_comparison} presents visualizations of retrievals by \emph{CriSp} and also compares them with those of prior methods.
    \item \Cref{sec:detailed_comparison_to_methods} provides a detailed quantitative comparison to state-of-the-art methods. We study generalization to unseen shoe models in \cref{subsec:study_generalization_to_unseen_shoe_models} and further detail the performance of methods on each unique shoe tread pattern in \cref{subsec:comparison_on_unique_tread_patterns}.
    \item \Cref{sec:training_prior_methods} elaborates on the training process of prior methods. We investigate the performance of fine-tuning these methods using masks to simulate partial prints in \cref{subsec:finetune_prior_methods_with_masks} and compare the performance of MCNCC\cite{kong2019cross} when using a reference database of shoeprints vs. tread depth maps in \cref{subsec:mcncc_with_print}.
    \item \Cref{sec:map} defines the mean average precision at K, which serves as a metric for evaluating and comparing methods.
    \item \Cref{sec:ground_truth_distribution_in_ref_db} analyses how the ground-truth shoe models are distributed within our reference database.
    \item  \Cref{sec:data_augmentation_details} provides detailed insights into our data augmentation technique.
    \item \Cref{sec:implementation_details} shares some implementation specifics of \emph{CriSp}.
\end{itemize}

\begin{table}[t]
    \centering
    \caption{Distribution of ground-truth shoe models from validation sets (val-FID and val-ShoeCase). We partition the ground-truth shoe models to assess generalization capabilities. In val-FID, there are 1152 shoe models, while val-ShoeCase comprises 16 shoe models. It's important to note that different shoe models may share tread patterns. Thus, we also distinguish between seen and unseen tread patterns during training. Val-FID encompasses 41 unique tread patterns, whereas val-ShoeCase contains 2 unique tread patterns.
    }
    \begin{tabular*}{\linewidth}{@{\extracolsep{\fill}}*{7}{c}} 
        \toprule
         & \multicolumn{3}{c}{shoe models} & \multicolumn{3}{c}{unique tread patters} \\
         &  seen& unseen &total & seen& unseen &total\\
         \midrule
         val-FID        &  229   &   923    & 1152  & 20 & 21 & 41\\
         val-ShoeCase   &  2    &   14       & 16    & 1 & 1 & 2\\
         \bottomrule
    \end{tabular*}
    \label{tab:real_crime_scene_data_label_statistics}
\end{table}

\begin{figure*}[t!]
\center
  \includegraphics[trim={0cm 7.4cm 17.1cm 2.2cm},clip,width=\linewidth]{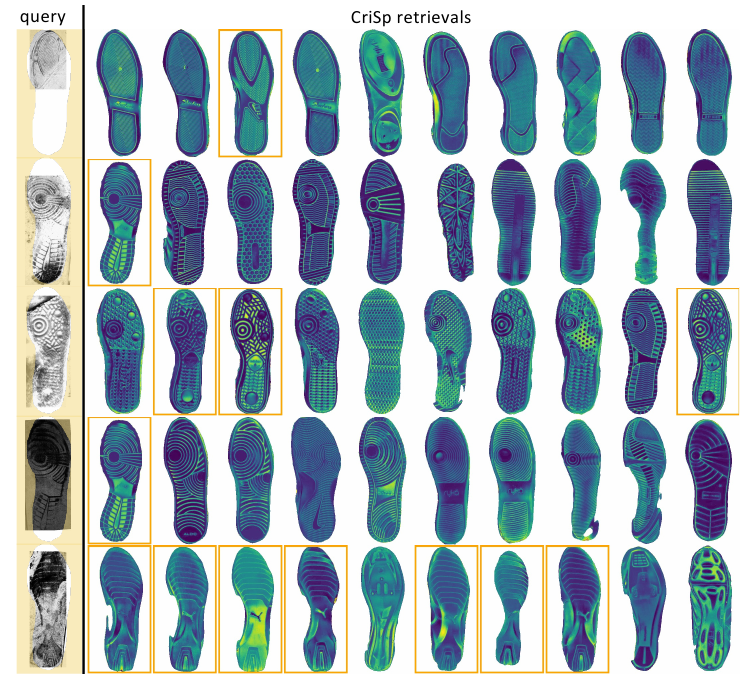}
  \caption{Qualitative results of \emph{CriSp} on val-FID. \emph{CriSp} retrieves positive matches early even with partially visible or severely degraded prints. 
  }
\label{fig:qual_fid_extra}
\end{figure*}

\begin{figure*}[t!]
\center
  \includegraphics[trim={0cm 9.55cm 17.1cm 2.2cm},clip,width=\linewidth]{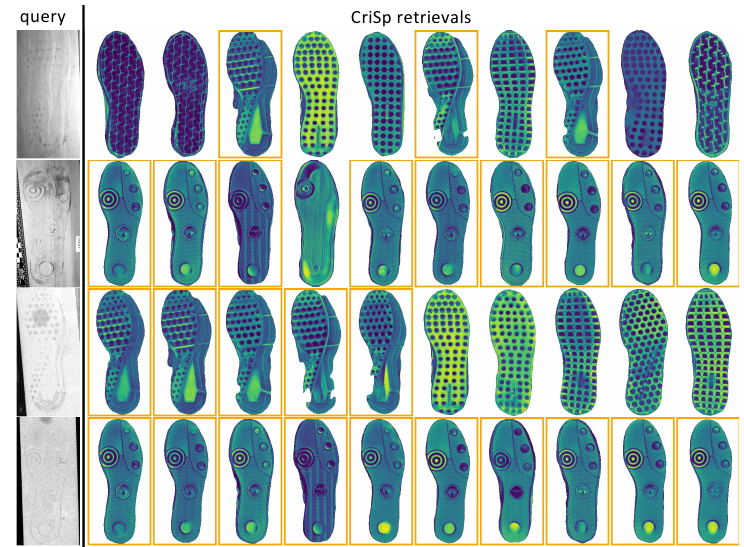}
  \caption{Qualitative results of \emph{CriSp} on val-ShoeCase. 
  We show the performance on prints from two different categories: blood prints (rows 1-2) and dust prints (rows 3-4). Despite the severe degradation present in the prints, \emph{CriSp} can retrieve positive matches early.
  }
\label{fig:qual_csafe_extra}
\end{figure*}

\begin{figure*}[t!]
\center
  \includegraphics[trim={0cm 5.3cm 19.9cm 2.2cm},clip,width=\linewidth]{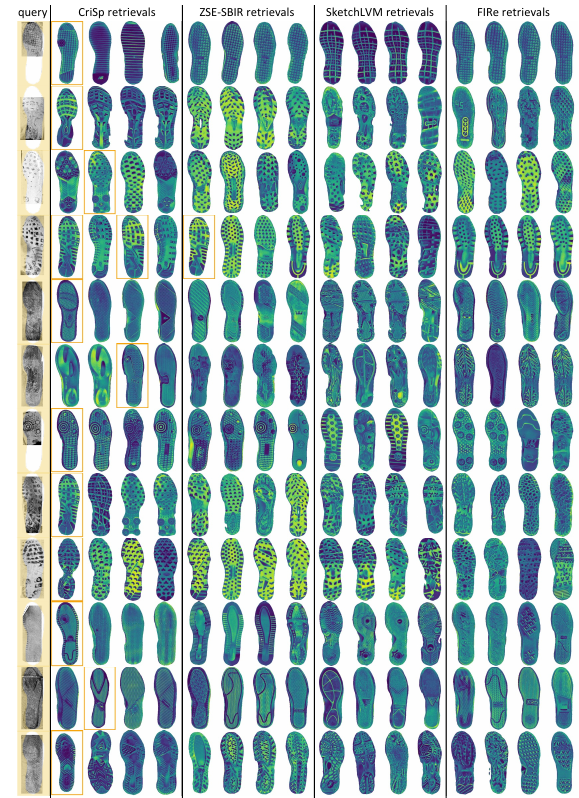}
  \caption{Qualitative comparison to state-of-the-art on val-FID. \emph{CriSp} outperforms prior methods by retrieving positive matches much earlier.
  }
\label{fig:qual_comparison_fid_extra}
\end{figure*}

\begin{figure*}[t!]
\center
  \includegraphics[trim={0cm 5.3cm 19.9cm 2.2cm},clip,width=\linewidth]{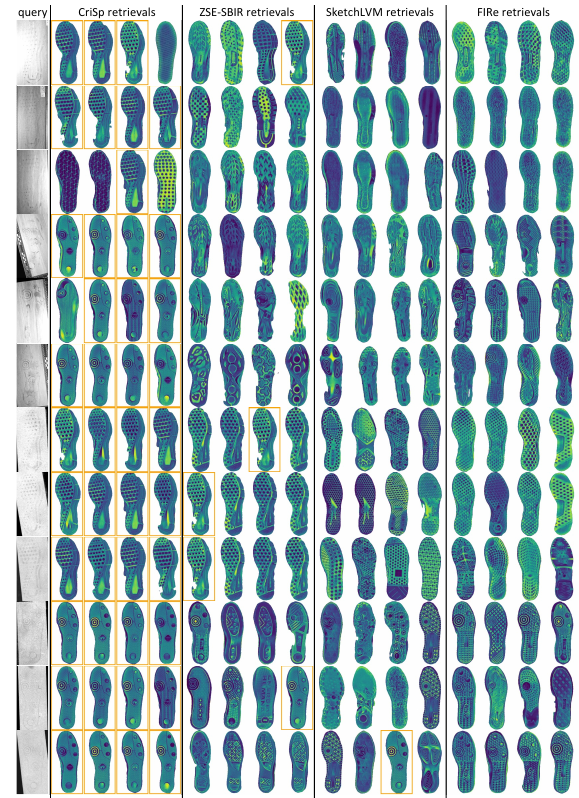}
  \caption{Qualitative comparison to state-of-the-art on val-ShoeCase. \emph{CriSp} outperforms prior methods by retrieving positive matches earlier, as evidenced by the top 6 rows displaying blood prints and the bottom 6 rows displaying dust prints.
  }
\label{fig:qual_comparison_csafe_extra}
\end{figure*}

\begin{table}[t!]
\caption{Benchmarking on validation sets to study generalization. 
We train prior methods on our dataset with our data augmentation technique. We compare the retrieval performance of methods using mAP@100. We categorize the shoeprints in the validation sets based on whether their corresponding tread patterns were seen during training or not. 
Note that we perform this study in terms of seen and unseen tread patterns instead of shoe models since multiple shoe models can share the same tread pattern. 
Notably, \emph{CriSp} demonstrates significantly superior performance to all prior methods on unseen tread patterns. However, ZSE-SBIR exhibits slightly better performance than \emph{CriSp} for seen tread patterns on val-ShoeCase.
}
\vspace{-3mm}
\setlength\tabcolsep{0pt} 
\begin{tabular*}{\linewidth}{@{\extracolsep{\fill}}*{1}{r}@{\extracolsep{2mm}}*{1}{l}@{\extracolsep{\fill}}*{4}{c}} 
\toprule
\multicolumn{2}{c}{\multirow{2}{*}{method}} & \multicolumn{2}{c}{val-FID} & \multicolumn{2}{c}{val-ShoeCase} \\
 & & seen & unseen & seen & unseen\\
\midrule
         IJCV'19 & MCNCC~\cite{kong2019cross} & 0.0002	& 0.0030	& 0.0000	& 0.0000 \\   %
         NeurIPS'20 & SupCon~\cite{supcon} &  0.0009	& 0.0000	& 0.0000	& 0.0000 \\  
         ICLR'21 & FIRe~\cite{fire} & 0.0671	& 0.0198	& 0.1103	& 0.2697 \\     
         CVPR'23 & SketchLVM~\cite{sketchLVM} &   0.0539	& 0.0270	& 0.0032	& 0.2770 \\ 
         CVPR'23 & ZSE-SBIR~\cite{zse-sbir} &  0.1659	& 0.1230	& \textbf{0.2653}	& 0.1350\\  
         & \textbf{CriSp} & \textbf{0.1749}	& \textbf{0.2309}	& 0.2495	& \textbf{0.4405} \\  
\bottomrule
\end{tabular*}
\vspace{-1mm}
\label{table:comparison_to_state_of_the_art_seen_unseen}
\end{table}

\begin{table}[t!]
    \centering
    \caption{We shoe mAP@100 for all unique tread patterns in val-FID. 
    \emph{CriSp} achieves the highest performance on 22 tread patterns, while ZSE-SBIR outperforms on 10 tread patterns. FIRe and SketchLVM exhibit the best performance on 1 tread pattern each.
    } 
    \setlength\tabcolsep{0pt} 
    \begin{tabular*}{\linewidth}{@{\extracolsep{\fill}}*{5}{c}} 
    \toprule
         tread pattern ID & FIRe &  SketchLVM  &  ZSE-SBIR  &  CriSp \\
          \midrule
000001	& 0.0000	& 0.0000	& \textbf{0.2583}	& 0.0027	\\
000003	& 0.0000	& 0.0000	& 0.0079	& \textbf{0.2333}	\\
000004	& 0.3108	& 0.0000	& 0.6137	& \textbf{0.6430}	\\
000005	& 0.1694	& 0.2083	& \textbf{0.5000}	& 0.4105	\\
000008	& 0.0025	& 0.0010	& 0.0034	& \textbf{0.0616}	\\
000009	& 0.0000	& 0.0000	& \textbf{0.0053}	& 0.0000	\\
000010	& 0.0000	& 0.0000	& 0.0250	& \textbf{0.5000}	\\
000011	& 0.0000	& 0.0000	& 0.4275	& \textbf{0.5060}	\\
000012	& 0.0067	& 0.0022	& 0.0713	& \textbf{0.0854}	\\
000013	& 0.0348	& 0.1145	& \textbf{0.2401}	& 0.0111	\\
000016	& 0.0000	& 0.0000	& 0.0563	& \textbf{0.0707}	\\
000017	& \textbf{0.1641}	& 0.0227	& 0.0105	& 0.0118	\\
000023	& 0.0000	& \textbf{0.3950}	& 0.0009	& 0.0148	\\
000032	& 0.0000	& 0.0000	& 0.0000	& 0.0000	\\
000033	& 0.0000	& 0.0000	& 0.2969	& \textbf{0.5711}	\\
000035	& 0.0000	& 0.0002	& \textbf{0.0027}	& 0.0000	\\
000045	& 0.0147	& 0.0000	& 0.0000	& \textbf{0.2500}	\\
000047	& 0.0000	& 0.0066	& 0.0000	& \textbf{0.0312}	\\
000053	& 0.0156	& 0.0000	& 0.0029	& \textbf{0.0427}	\\
000054	& 0.3148	& 0.0000	& \textbf{0.9444}	& 0.0265	\\
000055	& 0.0000	& 0.0000	& 0.0000	& \textbf{0.3258}	\\
000056	& 0.0000	& 0.0000	& 0.0000	& 0.0000	\\
000062	& 0.0000	& 0.0000	& 0.0000	& 0.0000	\\
000072	& 0.0018	& 0.0140	& 0.0054	& \textbf{0.0821}	\\
000074	& 0.0000	& 0.0000	& 0.0000	& \textbf{1.0000}	\\
000082	& 0.0000	& 0.0000	& 0.0000	& \textbf{0.0026}	\\
001040	& 0.0029	& 0.0000	& \textbf{0.0460}	& 0.0044	\\
001041	& 0.0000	& 0.0034	& 0.0027	& \textbf{0.2000}	\\
001044	& 0.2640	& 0.3070	& 0.4100	& \textbf{0.5091}	\\
001047	& 0.0000	& 0.0000	& 0.0000	& 0.0000	\\
001048	& 0.0000	& 0.0000	& \textbf{0.1036}	& 0.0000	\\
001049	& 0.0000	& 0.0100	& 0.0238	& \textbf{0.8333}	\\
001050	& 0.0038	& 0.1704	& 0.0437	& \textbf{0.3410}	\\
001058	& 0.0000	& 0.0000	& \textbf{0.3998}	& 0.0201	\\
001062	& 0.0000	& 0.0000	& 0.0000	& \textbf{0.4111}	\\
001064	& 0.0000	& 0.0000	& 0.0000	& \textbf{0.0006}	\\
001071	& 0.0000	& 0.0000	& 0.0000	& \textbf{0.0108}	\\
001076	& 0.0000	& 0.0000	& 0.0000	& 0.0000	\\
001079	& 0.0000	& 0.0000	& 0.0000	& 0.0000	\\
001088	& 0.0000	& 0.0000	& \textbf{0.5000}	& 0.0903	\\
001095	& 0.0000	& 0.0000	& 0.0000	& 0.0000	\\
\bottomrule
    \end{tabular*}
    
    \label{tab:map_per_shoe_model_val_FID}
\end{table}

\section{Qualitative Results of \emph{CriSp} and Comparison to State-of-the-art}
\label{sec:qual_analysis_and_comparison}

We display visualizations in this section. 
 \Cref{fig:qual_fid_extra} and \ref{fig:qual_csafe_extra} show the top 10 retrievals by \emph{CriSp} from crime-scene prints sourced from val-FID and val-ShoeCase, respectively.  These illustrations demonstrate \emph{CriSp}'s capability to retrieve positive matches even from severely degraded or partially visible crime-scene shoeprints. Furthermore, \Cref{fig:qual_comparison_fid_extra} and \ref{fig:qual_comparison_csafe_extra} provide a qualitative comparison between retrievals made by \emph{CriSp} and those of prior methods. Notably, \emph{CriSp} excels in matching patterns to corresponding regions on the tread, enabling it to retrieve positive matches early.

\section{Detailed Quantitative Comparison to State-of-the-art}
\label{sec:detailed_comparison_to_methods}

\subsection{Generalization to Unseen Shoe Models}
\label{subsec:study_generalization_to_unseen_shoe_models}

We compare our \emph{CriSp} to state-of-the-art methods to study generalization to unseen tread patterns. Note that we perform this study in terms of seen and unseen tread patterns instead of shoe models since multiple shoe models can share the same tread pattern. Our findings, detailed in \cref{table:comparison_to_state_of_the_art_seen_unseen}, demonstrate that \emph{CriSp} exhibits superior generalization to unseen tread patterns compared to prior methods.

\subsection{Comparison on Unique Tread Patterns}
\label{subsec:comparison_on_unique_tread_patterns}

We conduct a detailed analysis of \emph{CriSp} relative to prior methods on each unique tread pattern from val-FID. Recall that val-FID has 41 unique tread patterns among the 1152 ground-truth shoe models. \Cref{tab:map_per_shoe_model_val_FID} presents the comparison of methods based on mAP@100 for each tread pattern, where \emph{CriSp} exhibits superior performance in the majority of cases.

\section{Training State-of-the-art Methods}
\label{sec:training_prior_methods}

\subsection{Fine-tuning State-of-the-art Methods Using Simulated Crime-Scene Masks}
\label{subsec:finetune_prior_methods_with_masks}

When evaluating state-of-the-art methods, we train them on our dataset and apply our data augmentation to simulate crime-scene prints during training. Here, we compare the performance of related methods with and without using masks to simulate partial prints. Our findings are summarized in \cref{table:comparison_to_state_of_the_art_fid_masks}, demonstrating that \emph{CriSp} outperforms other methods in both settings.

\begin{table}[t!]
\caption{Benchmarking on real crime-scene prints from val-FID, we assess the impact of simulated partial print masks.
Using hit@100 and mAP@100 as metrics, we compare the performance of prior methods trained on our dataset with our data augmentation. The mAP@100 values reveal that prior methods tend to perform better when trained without masks simulating partial prints. \emph{CriSp} consistently achieves superior performance on both metrics, regardless of the presence of masks.}
\vspace{-3mm}
\setlength\tabcolsep{0pt} 
\begin{tabular*}{\linewidth}{@{\extracolsep{\fill}}*{1}{r}@{\extracolsep{2mm}}*{1}{l}@{\extracolsep{\fill}}*{4}{c}} 
\toprule
\multicolumn{2}{c}{\multirow{2}{*}{method}} & \multicolumn{2}{c}{w/o masks} & \multicolumn{2}{c}{w/ masks} \\
 & & hit@100 & mAP@100 & hit@100 & mAP@100\\
\midrule
         IJCV'19 & MCNCC~\cite{kong2019cross} &  0.0849 & 0.0018 & - & - \\   %
         NeurIPS'20 & SupCon~\cite{supcon} &  0.0755 & 0.0096 & 0.0849 & 0.0001 \\  
         ICLR'21 & FIRe~\cite{fire} & 0.2075 &  0.0398 & 0.0660 & 0.0030 \\       %
         CVPR'23 & SketchLVM~\cite{sketchLVM} &    0.1981 & 0.0384 & 0.2547 & 0.0445 \\ 
         CVPR'23 & ZSE-SBIR~\cite{zse-sbir} &  \textbf{0.4528} & 0.1412  & 0.4623 & 0.1358\\  
         & \textbf{Ours} &  \textbf{0.4528} & \textbf{0.1765} & \textbf{0.5472} & \textbf{0.2071}  \\  
\bottomrule
\end{tabular*}
\vspace{-1mm}
\label{table:comparison_to_state_of_the_art_fid_masks}
\end{table}


\subsection{Reference Database Configuration for MCNCC}
\label{subsec:mcncc_with_print}

When comparing MCNCC against a database of shoeprints, it yields a hit@100 of 0.0283 and mAP@K of 0.0008 on crime-scene shoeprints from val-FID. These metrics are notably lower compared to when using tread depths from the shoe database, where MCNCC achieves a hit@100 of 0.0849 and mAP@100 of 0.0018.




\section{Evaluation Metric - Mean Average Precision at K}
\label{sec:map}
Mean average precision at K (mAP@K) considers both the number of positive matches and their positions in the ranking list. It rewards the system’s ability to retrieve positive matches early. 
MAP@K is defined as follow:
\begin{align}
    \text{mAP@K} = \frac{1}{Q} \sum\limits_{q = 1}^{Q} AP@K_q
\end{align}
where $AP@K_q$ is the average precision at $K$ for query $q$. $AP@K$ is calculated as follows:
\begin{align}
    AP@K_q = \frac{1}{N} \sum\limits_{k = 1}^{K} Precision@k \times rel(k)
\end{align}
where $N$ is the total number of positive matches for a particular query.
Since we are only interested in the top $K$ retrievals, we limit $N$ to an upper bound of $K$.
$Precision(k)$ is the precision calculated at each position and is defined as $\frac{pos_k}{k}$ where $pos_k$ represents the number of positive matches in the top $k$ retrievals.
The final term, $rel(k)$, equals 1 if the item at position $k$ is a positive match and 0 otherwise. 

\section{Distribution of Shoe Models From Validation Sets in Reference Database}
\label{sec:ground_truth_distribution_in_ref_db}

Table~\ref{tab:real_crime_scene_data_label_statistics} provides insights into the distribution of ground-truth shoe models from our validation sets within the reference shoe database. Additionally, we present the count of distinct tread patterns that were either seen or unseen during training to facilitate comprehension. In \cref{subsec:study_generalization_to_unseen_shoe_models}, we assess the generalization performance of our model compared to state-of-the-art methods.



\section{Data augmentation to Simulate Crime-Scene Shoeprints}
\label{sec:data_augmentation_details}

Our data augmentation module $Aug$ simulates noisy and occluded crime-scene shoeprints from clean, fully-visible shoeprints. It introduces three types of degradation: occlusion, erasure, and noise. 
\begin{itemize}
    \item For occlusion, we simulate overlapping prints and quadrilaterals. Overlapping prints mimic the common occurrence of multiple shoeprints overlapping at a crime scene. We achieve this by randomly rotating and translating the predicted print and overlaying it onto itself. Quadrilaterals, resembling papers, rulers, or other marks, are added to simulate typical occlusions found in crime-scene shoeprint images.
    \item Erasure is incorporated to mimic the grainy nature of prints left at crime scenes. This involves selectively removing parts of the predicted shoeprint using either a Gaussian or Perlin distribution. Gaussian distribution is a standard choice for data augmentation, while Perlin noise provides a more nuanced representation of noise variations found in real images.
    \item Noise is added to represent background clutter. Gaussian or Perlin noise is overlaid on the predicted prints to simulate the clutter typically present in crime-scene images.
\end{itemize}
These degradations are applied dynamically during training, with each being optional.


\section{Implementation Details}
\label{sec:implementation_details}

We use a batch size of 4, where we randomly select 4 shoe models and then include two random instances per shoe model in each batch. During our experiments, training images of size 192 x 384 are encoded to a dimension of $H$=6 and $W$ = 12. 
We use an Adam optimizer with a learning rate of 0.1 and set $\tau=0.07$.

%
%

\end{document}